\documentclass{article}

\usepackage[preprint]{neurips_2026}

\usepackage[utf8]{inputenc} 
\usepackage[T1]{fontenc}    
\usepackage{hyperref}       
\usepackage{url}            
\usepackage{booktabs}       
\usepackage{amsfonts}       
\usepackage{nicefrac}       
\usepackage{microtype}      
\usepackage{xcolor}         
\usepackage{amsmath}
\usepackage{graphicx}
\usepackage{multirow}
\usepackage{makecell}

\newcolumntype{Y}{>{\raggedright\arraybackslash}X}

\title{3D MRI Image Pretraining via Controllable 2D Slice Navigation Task}

\author{%
  Yu Wang\\
  Beijing University of Posts and Telecommunications\\
  \texttt{wywork@bupt.edu.cn} \\
  \And
  Qingchao Chen\\
  Peking University\\
  \texttt{qingchao.chen@pku.edu.cn}
}

\begin{document}

\maketitle
\begin{abstract}
Self-supervised pretraining has become the mainstream approach for learning MRI representations from unlabeled scans. However, most existing objectives still treat each scan primarily as static aggregations of slices, patches or volumes. We ask whether there exists an intrinsic form of self-supervision signal that is different from reconstructing the masked patches, through transforming the 3D volumes into controllable 2D rendered sequences: by rendering slices at continuous positions, orientations, and scales, a 3D volume can be converted into dense video-action sequences whose controls are the action trajectories. We study this formulation with an action-conditioned pretraining objective, where a tokenizer encodes slice observations and a latent dynamics model predicts the evolution of latent features. Across representative anatomical and spatial downstream tasks, the proposed pretraining is evaluated against standard static-volume baselines, tokenizer-only pretraining, and dynamics variants without aligned actions. These results suggest that controllable MRI slice navigation provides a useful complementary pretraining interface for learning anatomical and spatial representations from large unlabeled MRI collections.
\end{abstract}

\section{Introduction}

Self-supervised pretraining has become a central strategy for learning 3D representations of MRI images from unlabeled scans \citep{zeng2024self, wald2025revisiting,9706678}. MRI is volumetric in 3D, \textit{but 2D slice views often mediate human interpretation}: radiologists tend to inspect MRI by scrolling stacks of 2D sliced images and switching among different 2D anatomical planes, rather than directly interpreting the 3D volumes \citep{crowe2018new, nakashima2016temporal}. \textit{This motivates us to ask a focused research question:} if a single 3D scan can be observed from many geometrically controlled 2D slice states, can their transitions serve as a intrinsic and useful self-supervised signal for pretraining?

\begin{figure}[t]
\centering
\includegraphics[width=\textwidth]{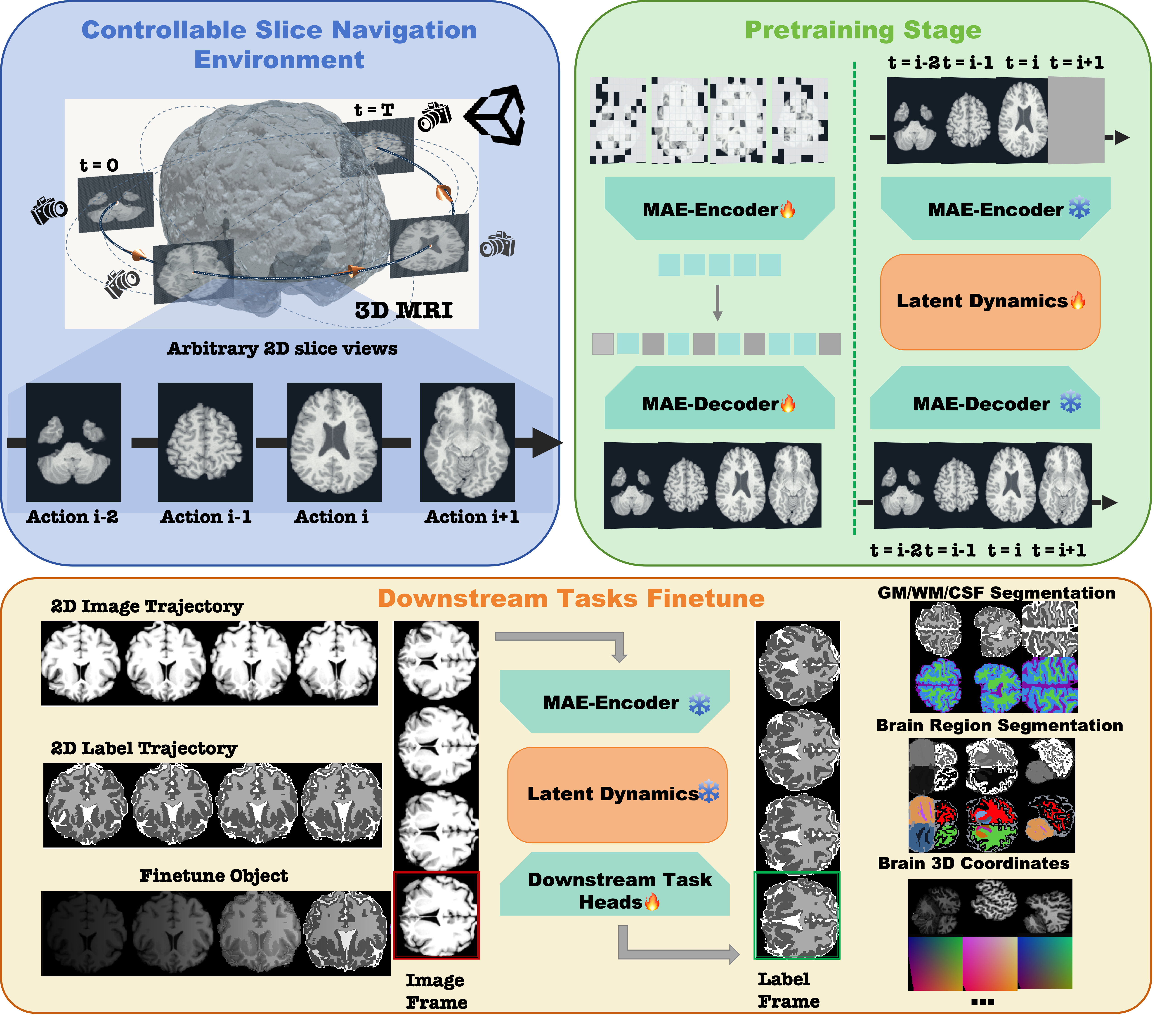}
\caption{Overview of controllable 2D slice navigation for 3D MRI pretraining. A fixed 3D MRI volume is re-rendered in Unity into arbitrary 2D slice videos, where each frame is generated by a recorded action controlling slice position, orientation, and scale. The resulting video-action sequences support tokenizer pretraining, action-conditioned latent dynamics pretraining, and downstream anatomical and spatial probes.}
\label{fig:overview}
\end{figure}

As illustrated in Figure~\ref{fig:overview}, a 3D MRI volume can be re-rendered as multiple 2D slice observations, spanning common anatomical planes-- axial, coronal, and sagittal-- as well as oblique views that are less frequently used in routine examinations. Typically, generating a 2D slice requires a specific set of action that defines its position, orientation and in-plane scaling factors. Therefore, by applying a sequence of parameters, a 3D volume can be transformed into a series of 2D sliceable image frames. These action parameters are not arbitrary labels or mere nuisance augmentations; rather, they specify how the same anatomical structure is observed \textit{from one viewing state to another}. Consequently, even an unlabeled volume can yield informative action-video sequences of slice transitions, containing and offering insights about spatial and structural relationships, that go beyond static, isolated views \citep{ZHANG2024103151, 10.1007/978-3-030-32245-8_60, bai2019self}. 

Although the action-video sequences contains richer information than the passive videos, effectively utilizing these action signals for MRI pretraining remains an open question. Recent natural video representation learning has shown that predicting latent content from context can yield meaningful representations without requiring the pixel-level reconstruction. Meanwhile, world-model methods extend the idea by learning action-conditioned latent dynamics, thereby incorporating the action/control parameters into the pretraining \citep{assran2025vjepa2, murlabadia2026vjepa2_1, maes_lelidec2026lewm, hafner2025dreamerv3, Hafner2025TrainingAgents}. Different from the above setup, the distinctive supervision in our pretraining may come from \textit{the known sequence of action parameters} in the 3D MRI rendering process. The process conveys the mechanism of rendering the slice observations given a sequence of action parameters, indicating how the slice plane moves, rotates, or changes scale. We therefore use the mechanism and formulate it in the following pretraining task: given a 2D MRI slice video and its recorded slice actions, the pretraining model learns to predict the next slice representation in the latent space \citep{acuaviva2025generation}.

We implement this formulation with a two-stage action-conditioned pretraining pipeline \citep{acuaviva2025generation}. First, a tokenizer is trained on rendered 2D MRI slices to produce spatial latent tokens. Second, with the tokenizer frozen, a latent dynamics model receives slice-token sequences together with their actions and learns to predict the latent representation of subsequent views. In this work, an action is simply a rendering control for a fixed volume: the position, orientation, and in-plane scale of the slice plane. It is not a clinical intervention, treatment decision, acquisition policy, or learned navigation policy. At finetune stage, we compare tokenizer features with dynamics features, where the frozen dynamics model provides action-conditioned context from the slice video for dense downstream prediction. Thus, the action is used to add controllable view-transition information.

We evaluate this formulation as a pretraining method, focusing on downstream probes that directly reflect its intended benefits. Brain-region segmentation and GM/WM/CSF segmentation test anatomical structure, while dense coordinate-field prediction tests whether the representation preserves 3D spatial organization. To isolate the role of action-conditioned dynamics, we compare tokenizer-only features, action-free dynamics, shuffled-action variants, and correctly action-conditioned dynamics. For segmentation, we further compare against static-volume and fixed-axis slice baselines \citep{wald2025revisiting, 9706678, MedSAM}, so that the proposed formulation is evaluated against both conventional 3D MRI and 2D slice-based alternatives. Together, these experiments provide evidence that controllable slice dynamics is a useful pretraining signal for MRI pretraining. Figure~\ref{fig:overview} summarizes the proposed controllable 2D slice-navigation formulation and the resulting pretraining pipeline.

Our contributions are threefold:
\begin{itemize}
    \item We define a controllable 2D slice-navigation formulation for 3D MRI pretraining, turning each unlabeled volume into many video-action sequences with recorded controls over slice position, orientation, and scale.
    \item We develop an action-conditioned pretraining pipeline that combines slice tokenization with latent dynamics, enabling dense downstream prediction from tokenizer-plus-dynamics features.
    \item We provide targeted evidence that action-conditioned slice dynamics is a useful signal for MRI pretraining, using anatomical segmentation, dense coordinate prediction, action-alignment diagnostics, and comparisons with static-volume and fixed-axis slice baselines.
\end{itemize}

\section{Related Work}

\paragraph{Self-supervised and spatially aware medical representation learning.}
Self-supervised pretraining originated as a response to the annotation bottleneck in medical imaging, and it has since become a standard route for learning transferable representations from unlabeled scans. \citep{zeng2024self, wald2025revisiting, 9706678} Recent reviews and foundation-model studies summarize a broad family of objectives, including reconstruction, contrastive learning, masked modeling, distillation, and hybrid pretraining across modalities \citep{zeng2024self, ZHANG2024102996}. In MRI and other volumetric settings, large-scale efforts such as LVM-Med, 3D masked autoencoding, and 3DINO further show that unlabeled medical volumes can support strong and transferable representations at scale \citep{xu2025generalizable,lang2023multispectral, NEURIPS2023_58cc11cd}. A complementary line of work originated from the observation that medical images often come with structured geometry: anatomy-oriented imaging planes and intrinsic spatial information can be exploited to pretrain representations that encode relative plane relations or physical crop location \citep{ZHANG2024103151, 10.1007/978-3-030-32245-8_60, bai2019self}. However, these methods still treat geometry mainly as a static target or a spatial proxy. In contrast, we ask whether the geometry of MRI observation itself can be turned into a dense pretraining signal, so that a 3D volume becomes a collection of aligned observation transitions rather than isolated static views.

\paragraph{World models and medical world modeling.}
World models originated in model-based reinforcement learning, where latent dynamics are learned from action-observation trajectories rather than from independent frames, and recent predictive representation learning has reinforced the value of forecasting latent content instead of reconstructing pixels \citep{battaglia2013simulation, zhou2021deep, hafner2025dreamerv3}. Recent medical work has begun to explore related formulations: CheXWorld studies image world modeling for radiographs, Medical World Model simulates disease evolution under clinical treatment decisions, and Cardiac Copilot uses a world model for echocardiographic probe guidance \citep{Yue_2025_CVPR, Yang_2025_ICCV, Jia_Cardiac_MICCAI2024}. These studies establish that world-model formulations can be useful in medicine, but their action spaces are tied to clinical intervention, disease progression, or acquisition policy. Our setting is narrower and more data-centric: we treat the slice position, orientation, and scale as retrospective observation controls over a fixed MRI volume, and we study whether action-conditioned latent dynamics can serve as a complementary self-supervised objective for learning reusable anatomical and spatial representations.

\section{Methods}
\subsection{Overall}

\begin{figure}[t]
\centering
\includegraphics[width=\textwidth]{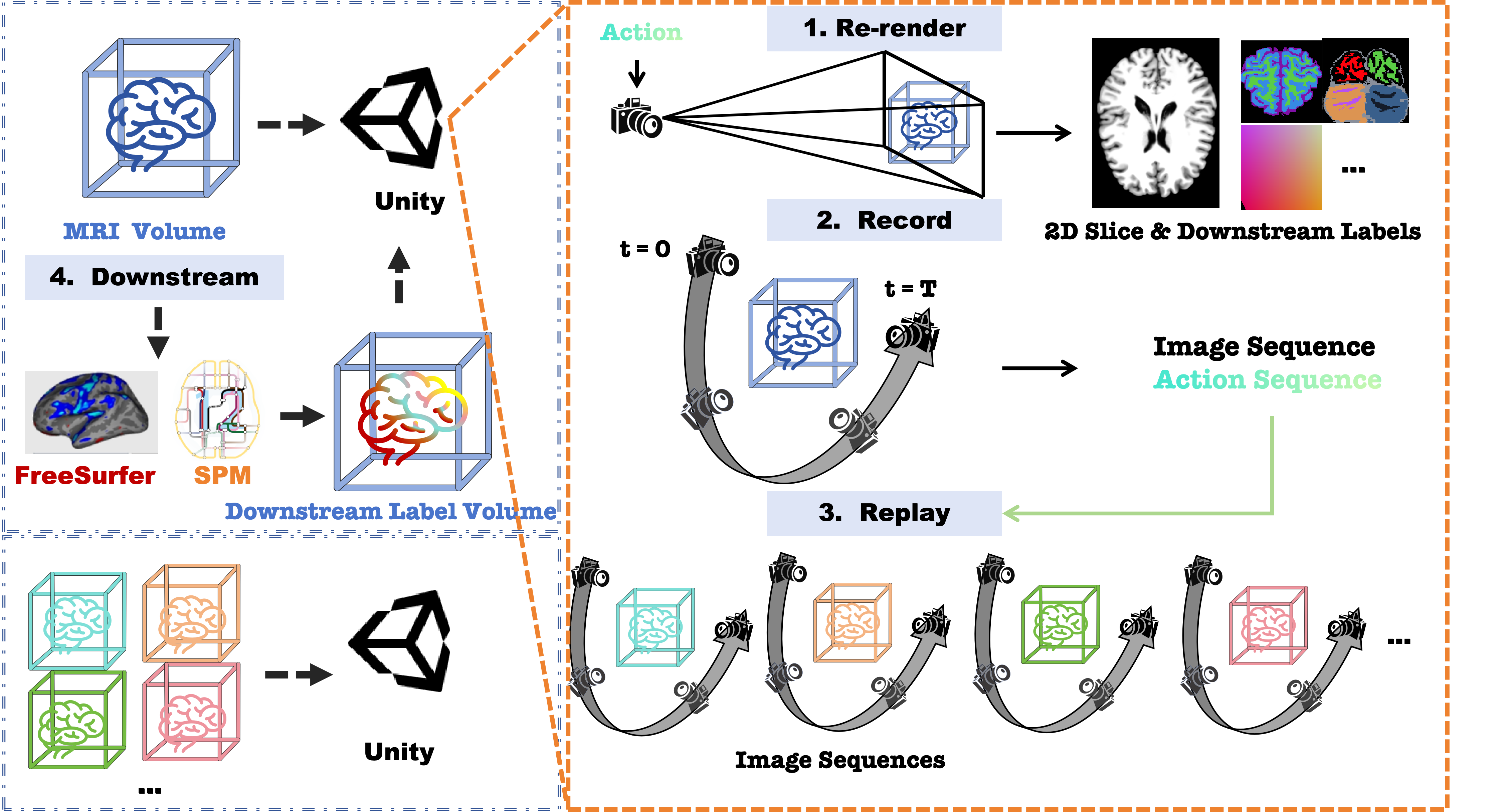}
\caption{Data preparation pipeline for 2D action-video and downstream supervision. (1) Re-render: each T1 MRI volume is loaded with SimpleITK and re-rendered in Unity into a 2D slice. (2) Record: the action-video is recorded manually in Unity. (3) Replay: the recorded action sequence is replayed on MRI volumes and downstream tasks label volumes.}
\label{fig:data_pipeline}
\end{figure}
Figure~\ref{fig:data_pipeline} gives an overview of data preparation pipeline. We load each 3D T1 MRI volume with SimpleITK \citep{10.3389/fninf.2013.00045} and re-render it in Unity into 2D slice video using manually recorded free-view action sequences. Our renderer is built on the open-source UnityVolumeRendering package \citep{lavikUnityVolumeRendering}. The resulting action-video data are used to pretrain a model, consisting of a tokenizer and an action-conditioned latent dynamics model. During finetuning, we train the task-specific downstream heads on top of the frozen tokenizer encoder and dynamics model.

\subsection{Controllable MRI slice trajectories}
Our innovative design and objective are to capture the \textit{temporal transitions of the rendered 2D slices}, and turn it into a self-supervised pretraining signal. Let $V_i \in \mathbb{R}^{H \times W \times D}$ denote a T1-weighted MRI volume from subject $i$. Instead of sampling only a fixed anatomical plane, we render $V_i$ through a controllable 2D slicing plane. For frame $t$, the Unity renderer takes a slice state $a_t$ and outputs:
\begin{equation}
    x_t = \mathcal{R}(V_i, a_t; \theta),
    \label{eq:slice_rendering}
\end{equation}
where $\mathcal{R}$ is the slice rendering operator, $\theta$ denotes fixed rendering parameters, and $x_t \in \mathbb{R}^{H \times W}$ is the rendered slice observation. The slice state of frame $t$ is represented as an 8-dimensional normalized control vector:
\begin{equation}
    a_t =
(p_x,p_y,p_z,r_{\mathrm{pitch}},r_{\mathrm{yaw}},r_{\mathrm{roll}},s_x,s_z)_t
    \in [-1,1]^8 .
    \label{eq:action_state}
\end{equation}

Here $(p_x,p_y,p_z)$ specifies the slice position, $(r_{\mathrm{pitch}},r_{\mathrm{yaw}},r_{\mathrm{roll}})$ specifies orientation, and $(s_x,s_z)$ specifies in-plane scaling factors. For each volume, we manually sample or replay a sequence of slice states and render the corresponding observations to get the action-video sequences:
\begin{equation}
    \tau_i = \{(x_t,a_t)\}_{t=0}^T.
    \label{eq:trajectory}
\end{equation}
The key property of this construction is alignment: $a_t$ is the slice state used to generate $x_t$, and consecutive pairs $(x_t,a_t),(x_{t+1},a_{t+1})$ describe how the observed anatomy changes under the varying slice states. The resulting pretraining dataset is
\begin{equation}
    \mathcal{D}_{\mathrm{pre}}
    =
    \{\tau_i\}_{i=1}^{N},
    \label{eq:pretraining_dataset}
\end{equation}
which provides dense action-observation supervision without manual labels.

To produce frame-matched targets $y_i$, we replay the same state sequence using the actions  in $\tau_i$ and the downstream target volumes $Y_i$:
\begin{equation}
    y_i = \mathcal{R}(Y_i,a_t;\theta).
    \label{eq:label_rendering}
\end{equation}

\subsection{World-model pretraining objective}
As shown in Figure \ref{fig:overview}, we adopted an encoder-decoder architecture and proposed a novel two-stage pretraining pipeline using the MRI 2D action-video pairs. The \textit{first} stage learns a tokenizer that maps individual slice observations to latent tokens. The \textit{second} stage freezes the tokenizer and trains an action-conditioned latent dynamics model on the action-video sequences. This decomposition separates static anatomical encoding from transition learning: the tokenizer provides a compact latent feature space \textit{while} the dynamics model learns how to \textit{evolve} latent features of anatomical views by predicting the next latent frame.

For a slice observation $x_t$, let $P(x_t) \in \mathbb{R}^{N_p \times d_p}$ denote its patchified representation and let $m_t \in \{0,1\}^{N_p}$ be a binary mask over patches. The tokenizer $ E_\psi$ outputs the latent token $z_t \in \mathbb{R}^{L \times d}$ and the decoder $G_\psi$ predicts the estimated patches $\hat{P}_t$ as follows: 
\begin{equation}
    z_t = E_\psi(P(x_t),m_t),
    \qquad
    \hat{P}_t = G_\psi(z_t,m_t).
    \label{eq:tokenizer}
\end{equation}
The tokenizer is trained with masked patch reconstruction:
\begin{equation}
    \mathcal{L}_{\mathrm{mae}}
    =
    \mathbb{E}_{x_t,m_t}
    \left[
    \frac{1}{\sum_{j=1}^{N_p} m_{t,j}}
    \sum_{j=1}^{N_p}
    m_{t,j}
    \left\|
        \hat{P}_{t,j} - P(x_t)_j
    \right\|_2^2
    \right].
    \label{eq:mae_loss}
\end{equation}
When perceptual reconstruction is enabled, we add a perceptual term between the reconstructed and target slice:
\begin{equation}
    \mathcal{L}_{\mathrm{token}}
    =
    \mathcal{L}_{\mathrm{mae}}
    +
    \lambda_{\mathrm{perc}}
    \mathcal{L}_{\mathrm{perc}}
    \bigl(
        \operatorname{Unpatchify}(\hat{P}_t), x_t
    \bigr).
    \label{eq:tokenizer_loss}
\end{equation}
The $\operatorname{Unpatchify}$ operator is to reconstruct the estimated image $\hat{x_t}$ from the estimated patches $\hat{P_t}$ with the correct positions.

After tokenizer pretraining, we freeze the tokenizer encoder $E_\psi$ and train the dynamics model only in the tokenizer latent space. For a slice-action trajectory $\tau_i=\{(x_t,a_t)\}_{t=1}^{T}$, each frame is encoded as
\begin{equation}
    z_t = \operatorname{Pack}(E_\psi(x_t)),
    \qquad
    z_t \in \mathbb{R}^{S \times d_z},
    \label{eq:dynamics_latent}
\end{equation}
where $\operatorname{Pack}(\cdot)$ denotes the spatial packing operation used by the dynamics model. The clean target $z_t$ is therefore the frozen-tokenizer representation of the rendered slice $x_t$; no manual label or image-space reconstruction target is used in this stage.

We train the dynamics model with a shortcut-forcing denoising objective. For each time step, a shortcut signal $\sigma_t \in [0,1]$ is sampled and the clean latent is mixed with Gaussian noise:
\begin{equation}
    \tilde{z}_t^{\sigma}
    =
    (1-\sigma_t)\epsilon_t + \sigma_t z_t,
    \qquad
    \epsilon_t \sim \mathcal{N}(0,I).
    \label{eq:shortcut_noise}
\end{equation}
Here $\sigma_t$ controls the denoising level: $\sigma_t=0$ corresponds to pure noise and $\sigma_t=1$ corresponds to the clean tokenizer latent. The shortcut step embedding $\Delta_t$ follows the Dreamer4-style shortcut objective and specifies the denoising step size, rather than a temporal offset in the MRI trajectory.

The dynamics model $F_\phi$ is causal over the slice sequence. At time $t$, it receives the corrupted latent context, the recorded slice states up to $t$, the shortcut conditioning variables, and predicts the clean latent $\hat{z}_t$ of the current rendered view
\begin{equation}
    \hat{z}_t
    =
    F_\phi
    \left(
        \tilde{z}_{\leq t}^{\sigma},
        a_{\leq t},
        \sigma_{\leq t},
        \Delta_{\leq t}
    \right).
    \label{eq:dynamics_prediction}
\end{equation}

The empirical dynamics loss is
\begin{equation}
    \mathcal{L}_{\mathrm{emp}}
    =
    \mathbb{E}_{\tau,t,\sigma}
    \left[
        w(\sigma_t)
        \left\|
            \hat{z}_t - z_t
        \right\|_2^2
    \right],
    \label{eq:dynamics_empirical_loss}
\end{equation}
where $w(\sigma_t)$ is the shortcut weighting. When shortcut bootstrapping is enabled, we used the dynamic loss $\mathcal{L}_{\mathrm{dyn}}$ by adding the standard shortcut self-consistency term $\mathcal{L}_{\mathrm{sc}}$ from the Dreamer4 \citep{acuaviva2025generation} objective:
\begin{equation}
    \mathcal{L}_{\mathrm{dyn}}
    =
    \mathcal{L}_{\mathrm{emp}}
    +
    \lambda_{\mathrm{sc}}\mathcal{L}_{\mathrm{sc}} .
    \label{eq:dynamics_total_loss}
\end{equation}
This self-consistency term uses model predictions only and is shared by all dynamics variants. The only external signal consumed by the dynamics stage is the recorded slice-action sequence. 

\subsection{Transfer to downstream tasks}
\label{sec:downstream_transfer}

We evaluate whether the pretrained representation transfers to dense anatomical and spatial prediction. Given a $K$-frame slice clip $c_t$ ending at time $t$,
\begin{equation}
    c_t = \{(x_{t-K+1},a_{t-K+1}),\ldots,(x_t,a_t)\},
    \label{eq:downstream_clip}
\end{equation}
the frozen pretrained backbone produces a representation $h_t=\rho(c_t)$, and a lightweight task-specific head predicts the target for the final slice:
\begin{equation}
    \hat{y}_t = g_\omega(h_t).
    \label{eq:downstream_prediction}
\end{equation}
Here $\rho$ denotes tokenizer-plus-dynamics features. The pretrained tokenizer and dynamics model are frozen, so downstream performance reflects the transferability of the learned representation.
\section{Experiments}

Our experiments evaluate whether controllable MRI slice navigation provides a useful pretraining signal for anatomical and spatial representation learning. The experiments are organized around four questions. First, does the pretrained representation improve anatomical dense prediction compared with baselines and other methods? Second, how much slice-video context is needed, and what trade-off does context length create between accuracy, inference time, and memory cost? Third, does the dynamics model actually use the recorded slice actions during pretraining? Fourth, does the learned representation preserve 3D spatial structure beyond local slice appearance? 

\subsection{Experimental setup}

We use 184 T1-weighted MRI volumes from the HCP2 dataset \citep{VANESSEN20122222}. All experiments use subject-level splits with a fixed random seed: 148 subjects for training, 18 for validation, and 18 for testing. The test subjects are held out from both self-supervised pretraining and downstream training. Volumes are loaded with a SimpleITK-based NIfTI pipeline \citep{10.3389/fninf.2013.00045}; the Unity rendering and target replay pipeline is summarized in Figure~\ref{fig:data_pipeline}.

Unless otherwise specified, rendered slice frames are resized to $256\times256$ and paired with 8-dimensional slice states encoding position, orientation, and in-plane scale. The tokenizer is pretrained on grayscale slice videos with patch size 16. The dynamics model is trained on strided slice-action sequences with frame stride 4. For downstream transfer, we train lightweight prediction heads on frozen pretrained features, sweep clip lengths $K\in\{2,4,8,16\}$, and select models by validation performance.

We evaluate three dense probes. Brain-region segmentation uses FreeSurfer-derived parcellation labels replayed with the same slice states, and evaluates whether the representation captures anatomical region structure \citep{fischl2012freesurfer}. GM/WM/CSF segmentation uses SPM-derived tissue labels and evaluates tissue-level anatomical information \citep{ashburner2009computational}. Dense 3D coordinate-field prediction uses Unity-recorded coordinates for valid slice pixels and evaluates whether the representation preserves spatial organization in the underlying 3D volume. Segmentation performance is reported with Dice and mIoU, while coordinate prediction is reported with valid-mask MAE and per-axis MAE. Task-specific losses, valid-pixel handling, and metric definitions are provided in the supplementary material.

We compare tokenizer-only pretraining, no-action dynamics, and action-conditioned dynamics. For segmentation, we additionally include static-volume baselines: supervised UNETR \citep{9706678} and S3D-style 3D self-supervised pretraining \citep{wald2025revisiting}. These baselines test whether controllable slice-trajectory pretraining provides value beyond conventional 3D volume learning and fixed-axis 2D slice segmentation.

\subsection{Main downstream results}
\begin{table}[t]
\centering
\caption{Few-epoch downstream adaptation. We report validation Dice for segmentation and validation MAE for coordinate prediction after 1, 3, and 5 fine-tuning epochs. }
\label{tab:method_comparison}
\renewcommand{\arraystretch}{1.08}
\begin{tabular}{@{}clccc@{}}
\toprule
Epoch
& Method
& Region Dice $\uparrow$
& Tissue Dice $\uparrow$
& Coord. MAE $\downarrow$ \\
\midrule

\multirow{3}{*}{1}
& Tokenizer-only & 0.5058 & 0.8359 & 0.0215 \\
& Tokenizer + no-action dyn. & 0.5131 & 0.8384 & 0.0181 \\
& Tokenizer + action-cond. dyn. & \textbf{0.5157} & \textbf{0.8385} & \textbf{0.0177} \\
\addlinespace[2pt]
\midrule

\multirow{3}{*}{3}
& Tokenizer-only & 0.5133 & 0.8461 & 0.0149 \\
& Tokenizer + no-action dyn. & \textbf{0.5475} & 0.8483 & 0.0137 \\
& Tokenizer + action-cond. dyn. & 0.5471 & \textbf{0.8491} & \textbf{0.0134} \\
\addlinespace[2pt]
\midrule

\multirow{3}{*}{5}
& Tokenizer-only & 0.5449 & 0.8538 & 0.0117 \\
& Tokenizer + no-action dyn. & 0.5721 & 0.8544 & 0.0090 \\
& Tokenizer + action-cond. dyn. & \textbf{0.5736} & \textbf{0.8552} & \textbf{0.0090} \\
\bottomrule
\end{tabular}

\vspace{2pt}

\end{table}

\begin{table}[t]
\centering
\caption{Segmentation results for static-volume baselines and controllable 2D slice navigation pretraining task.}
\label{tab:seg_detail}
\begin{tabular}{@{}lcccc@{}}
\toprule
Method
& Region Dice $\uparrow$
& Region mIoU $\uparrow$
& Tissue Dice $\uparrow$
& Tissue mIoU $\uparrow$ \\
\midrule
\multicolumn{5}{l}{\textit{Static 3D volume baselines}} \\
3D UNETR \citep{9706678} & 0.6772 & 0.5271 & 0.8802 & 0.7932 \\
S3D \citep{wald2025revisiting}  & 0.7674 & 0.6343 & 0.8951 & 0.8166 \\
\midrule
\multicolumn{5}{l}{\textit{Our 2D Slice Navigation pretraining task}} \\
Tokenizer-only & 0.5450 & 0.4239 & 0.8538 & 0.7536 \\
Tokenizer + no-action dyn. & 0.5721 & 0.4474 & 0.8544 & 0.7540 \\
Tokenizer + action-cond. dyn. & 0.5736 & 0.4491 & 0.8552 & 0.7552 \\
\bottomrule
\end{tabular}
\end{table}

As illustrated in Table~\ref{tab:method_comparison}, the action-conditioned dynamics model consistently improves the few-epoch adaptation at the 1/3/5 epoch setting over the tokenizer-only pretraining and the tokenizer plus no-action dynamics model pretraining. These results support the main hypothesis that correct actions provide useful transition supervision beyond static tokenizer pretraining. 

\begin{figure}[t]
\centering
\includegraphics[width=\textwidth]{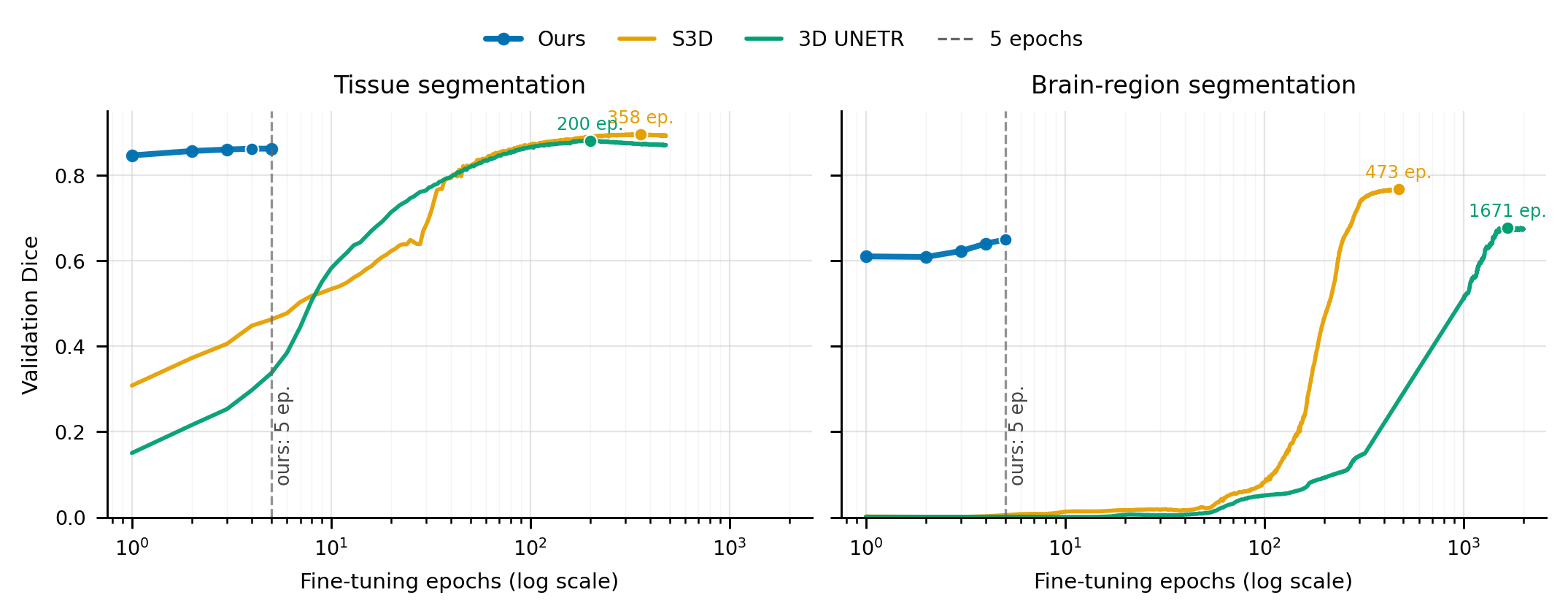}
\caption{Fine-tuning efficiency on tissue and brain-region segmentation.
Our action-conditioned world-model representation reaches strong downstream performance within 5 epochs, whereas static 3D baselines require hundreds to thousands of fine-tuning epochs.
For S3D, only supervised fine-tuning epochs are counted; its masked 3D pretraining epochs are not included.
Curves are parsed from the downloaded training logs.}
\label{fig:time_comparation}
\end{figure}

\begin{figure}[t]
\centering
\includegraphics[width=\textwidth]{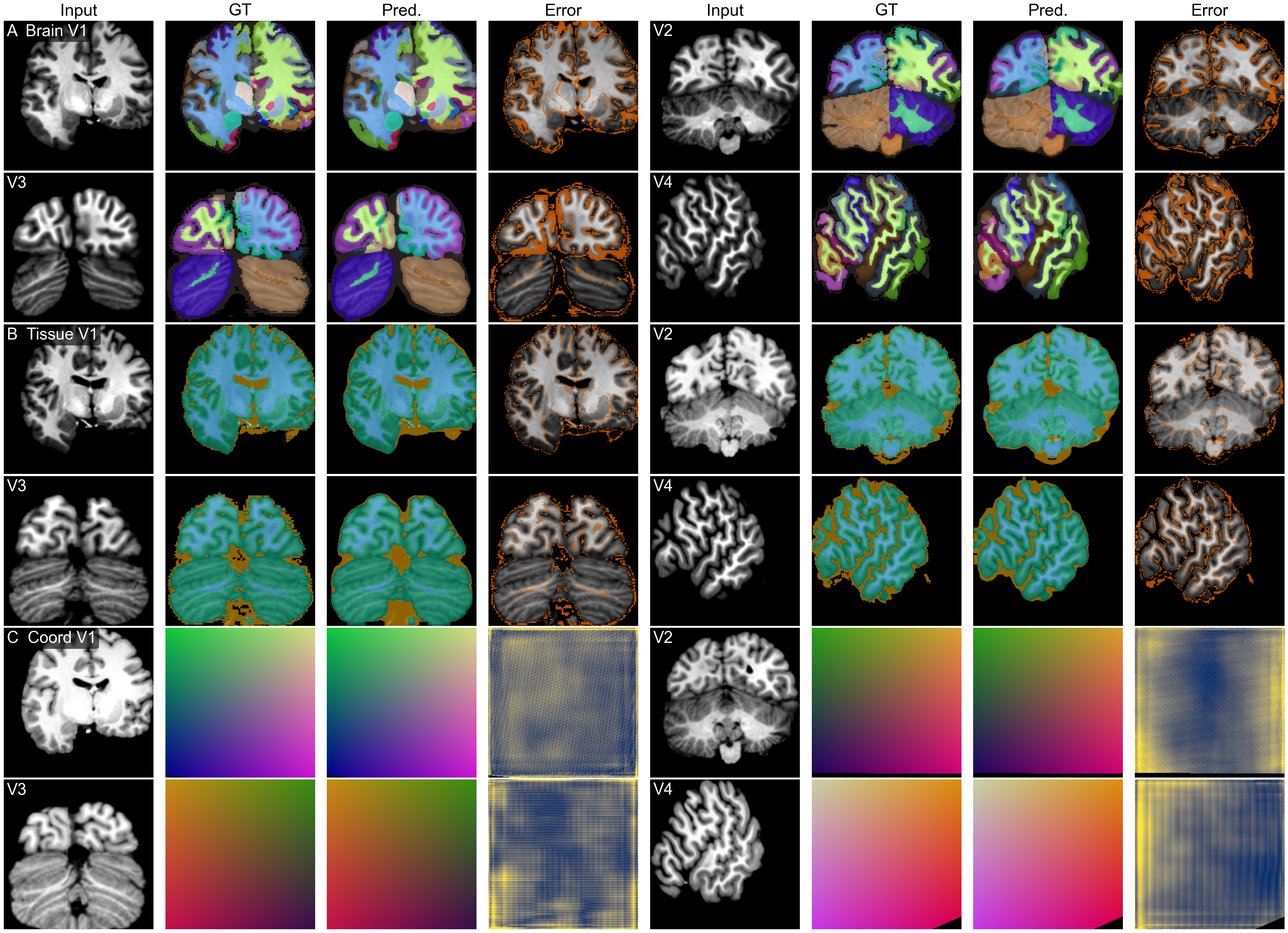}
\caption{The visualized results of the brain region segmentation, the tissue segmentation, and the dense 3D coordinates understanding tasks.}
\label{fig:replay_targets}
\end{figure}

Table~\ref{tab:seg_detail} further compares the segmentation results with the supervised static-volume segmentation method 3D UNETR \citep{9706678} and the self-supervised and fine-tuned static-volume segmentation method S3D \citep{wald2025revisiting}. The proposed slice-navigation pretraining does not yet match fully optimized 3D baselines such as UNETR and S3D on segmentation accuracy. As shown in Figure~\ref{fig:time_comparation}, this gap may be expected under the current evaluation protocol: our downstream adaptation trains lightweight heads for only 5 epochs with frozen pretrained modules, whereas the static 3D baselines are optimized for substantially longer schedules, with brain-region UNETR trained for 2000 epochs and S3D fine-tuned for 480 epochs. The key observation is that controllable slice dynamics already yields measurable gains under a short adaptation budget, suggesting that action-conditioned pretraining provides a useful signal for efficient downstream adaptation. The visualized results are shown in Figure~\ref{fig:replay_targets}

\subsection{Effect of slice-video context length}
\begin{table}[t]
\centering
\caption{Effect of slice-video context length for action-conditioned dynamics. Accuracy is reported on the test set. Latency and memory are measured on a single NVIDIA A800 with batch size 1 and averaged across the three downstream probes.}
\label{tab:sequence_length}
\begin{tabular}{@{}c ccc cc@{}}
\toprule
\multirow{2}{*}{$L$}
& \multicolumn{3}{c}{Accuracy}
& \multicolumn{2}{c}{Cost} \\
\cmidrule(lr){2-4}\cmidrule(l){5-6}
& Region Dice $\uparrow$
& Tissue Dice $\uparrow$
& Coord. MAE $\downarrow$
& Lat. (ms)
& Mem. (GB) \\
\midrule
2  & 0.5927 & 0.8484 & 0.0100 & 16.87 & 0.380 \\
4  & 0.5919 & 0.8493 & 0.0099 & 16.77 & 0.381 \\
8  & 0.5899 & 0.8482 & 0.0093 & 16.65 & 0.382 \\
16 & \textbf{0.6749} & \textbf{0.8542} & \textbf{0.0085} & 16.92 & 0.433 \\
\bottomrule
\end{tabular}
\end{table}

As shown in Table~\ref{tab:sequence_length}, we change $L$, the number of context frames, to observe the performance of downstream tasks. The short context frames already yield competitive tissue segmentation and coordinate prediction performance, but the strongest overall results are obtained at $L=16$. Increasing $L$ from 2 to 16 improves region Dice from 0.5927 to 0.6749, tissue Dice from 0.8484 to 0.8542, and coordinate MAE from 0.0100 to 0.0085. The improvement is especially pronounced, suggesting that longer slice trajectories provide additional anatomical and structural information for downstream tasks. Besides, the computational overhead is modest in this setting. Latency remains nearly unchanged across context lengths, ranging from 16.65 ms to 16.92 ms on a single A800, while memory increases from 0.380 GB at $L=2$ to 0.433 GB at $L=16$. 

\subsection{Action-use diagnostics}
\begin{figure}[t]
\centering
\includegraphics[width=\textwidth]{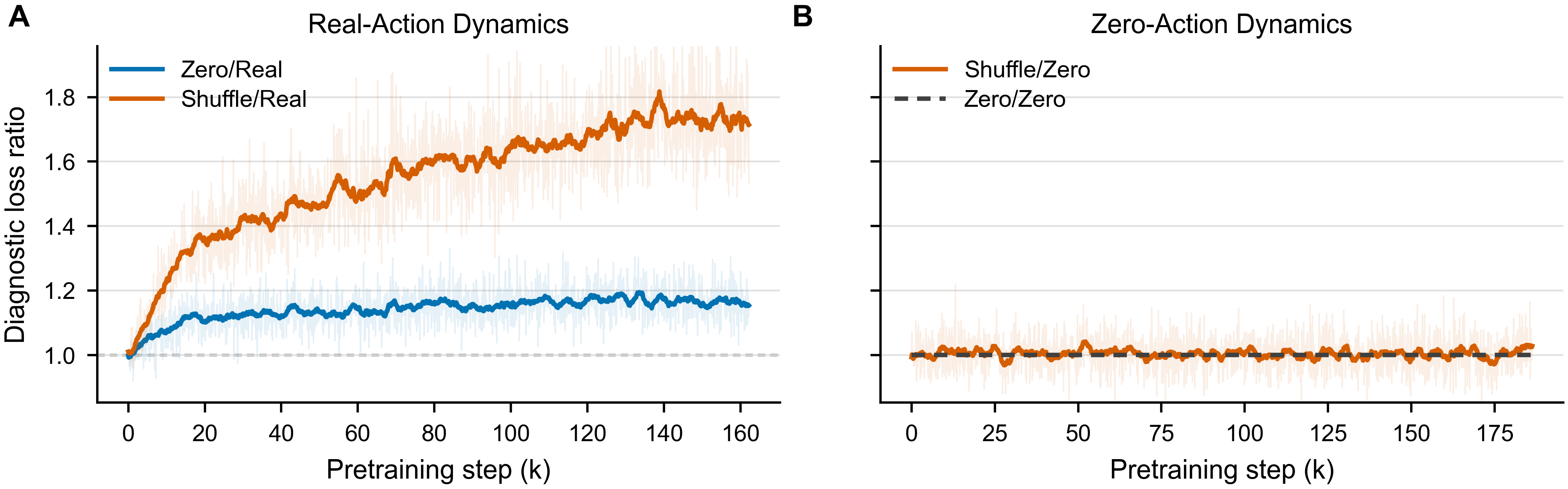}
\caption{Action-use diagnostics during dynamics pretraining. In the Real-Action Dynamics model, we report diagnostic loss ratios under Zero and Shuffle actions relative to Real actions, denoted as Zero/Real and Shuffle/Real. In the Zero-Action Dynamics model, we use Zero actions as the denominator and report Shuffle/Zero and Zero/Zero. Curves show a centered moving average with translucent raw traces.}
\label{fig:action_ratio}
\end{figure}

We further ask whether the dynamics model actually uses the slice actions during the pretraining. To test this, we replace only the action used by the dynamics model every single training step, but calculate gradients only with the correct action. For the real-action model, we compare three inputs: the recorded actions, zero actions, and shuffled actions. If the model relies on correct action-observation alignment, replacing the recorded actions should increase the latent prediction loss. As a control, we apply the same test to a zero-action model, which is trained without informative actions.

Figure~\ref{fig:action_ratio} reports the loss ratios during training. In the real-action model, zero or shuffled actions produce higher prediction loss than the recorded actions, showing that the model learns to use the slice actions. In the zero-action model, the ratios remain much closer to one, indicating weaker dependence on the action input. These results support that the benefit of our dynamics pretraining comes from correct dynamics.

\section{Conclusion and Limitations}

We studied whether controllable MRI slice navigation can serve as a self-supervised pretraining. By viewing a 3D MRI volume as an environment that can be observed through controllable slice states, each unlabeled scan can generate dense action-slice trajectories. We instantiated this idea with a world-model pretraining objective, where a tokenizer learns compact slice representations and an action-conditioned dynamics model learns latent transitions between anatomical views. Through downstream anatomical and spatial probes, together with action ablations, our results provide initial evidence that controllable slice dynamics is a useful signal for MRI representation learning.

This work has several limitations. First, our evaluation is intentionally focused: we use a limited set of dense downstream probes rather than a broad benchmark across MRI modalities, organs, and clinical tasks. Second, the current experiments are centered on HCP2 T1 MRI, so the generality of the formulation to other contrasts, scanners, pathologies, and body regions remains open. Third, our current environment uses slice observation controls rather than richer interactive or physical actions.

Future work can extend this direction in several ways. A first step is to scale pretraining to larger and more diverse MRI collections, including multi-contrast and multi-site data. A second direction is to study richer controllable action spaces beyond geometric slice motion alone. Finally, downstream evaluation should be expanded to broader anatomical and clinical tasks, including low-label and cross-domain transfer settings. We hope this paper is an initial step toward using controllable 2D slice actions as  powerful signals for 3D MRI image pretraining.

\newpage
\bibliographystyle{plain}
\bibliography{References}
%
%
%
%


\end{document}